\documentclass[conference]{IEEEtran}
\IEEEoverridecommandlockouts
\usepackage{amsmath}
\usepackage{graphicx}
\usepackage{booktabs}
\usepackage{amssymb}
\usepackage{latexsym} 
\usepackage{array}    
\usepackage{multirow}
\usepackage{subcaption}
\usepackage{algorithm}
\usepackage[noend]{algpseudocode}
\usepackage[font={small,bf}]{caption}
\usepackage{threeparttable}
\usepackage{paralist}   
\usepackage{xspace}
\usepackage[dvipsnames]{xcolor}  
\usepackage{soul}
\usepackage{adjustbox}
\usepackage{wasysym}
\usepackage{rotating}   
\usepackage{listings}
\usepackage{pifont}
\usepackage{framed}

\usepackage{url}            

\definecolor{wheat1}{rgb}{1.000000,0.905882,0.729412}
\definecolor{LightGray}{rgb}{0.827451,0.827451,0.827451}

\newcolumntype{a}{>{\columncolor{wheat1}}l}

\definecolor{mygreen}{rgb}{0,0.6,0}
\definecolor{mygray}{rgb}{0.5,0.5,0.5}
\definecolor{mymauve}{rgb}{0.58,0,0.82}
\definecolor{darkblue}{rgb}{0.0,0.0,0.6}
\definecolor{maroon}{RGB}{102, 0, 0}
\definecolor{Maroon}{cmyk}{0,0.87,0.68,0.32}
\definecolor{darkred}{RGB}{139, 0, 0}
\definecolor{forestgreen}{RGB}{34, 139, 34}

\lstdefinelanguage{XML}
{
  basicstyle=\ttfamily\small,   
  morestring=[b]",
  moredelim=[s][\color{darkblue}]{<}{\ },
  moredelim=[s][\color{darkblue}]{</}{>},
  moredelim=[l][\color{darkblue}]{/>},
  moredelim=[l][\color{darkblue}]{>},
  morecomment=[s]{<?}{?>},
  morecomment=[s]{<!--}{-->},
  stringstyle=\color{darkred},
  identifierstyle=\color{mymauve}
}

\lstdefinestyle{customJava}{
  breaklines=true,
  keepspaces=true,
  frame=single,
  language=Java,
  showstringspaces=false,
  basicstyle=\footnotesize\ttfamily,
  keywordstyle=\color{blue},
  otherkeywords={+, getIntent},
  numbers=left,
  numbersep=5pt,
  numberstyle=\scriptsize\color{black},
  rulecolor=\color{black},
  stepnumber=1,
  tabsize=2,
  commentstyle=\itshape\color{green!40!black},
  stringstyle=\color{orange},
  emph=[1]  
  {
        do,
        try,
        new,
        catch,
        while,
        SecProvider,
        SecReceiver,
        SecService,
        SecActivity,
        SecSink,
  },
  emphstyle=[1]{\color{darkred}},
  emph=[2]  
  {
        @Override,
  },
  emphstyle=[2]{\color{purple!40!black}},
  belowskip=-1em, 
}

\newif\ifANNOYMIZE
\ANNOYMIZEfalse

\newif\ifACM
\ACMfalse 

\ifACM
\fi

\ifACM
\newcommand{\myfig}{Figure.\xspace}
\else
\newcommand{\myfig}{Fig.\xspace}
\fi

\ifACM

\else

\fi

\ifACM

\else

\fi

\newcommand{\name}{iExam\xspace}

\def\BibTeX{{\rm B\kern-.05em{\sc i\kern-.025em b}\kern-.08em
    T\kern-.1667em\lower.7ex\hbox{E}\kern-.125emX}}
\begin{document}

\title{Effective Online Exam Proctoring by Combining Lightweight Face Detection and Deep Recognition}

\ifANNOYMIZE
\author{\IEEEauthorblockN{Anonymous ICME Submission}
}
\else
\IEEEspecialpapernotice{Website: \texttt{https://vprlab.github.io/iexam}\\Code: \texttt{https://github.com/VPRLab/iExam}}

\author{
\IEEEauthorblockN{Xu Yang$^{\dagger}$\thanks{$^{\dagger}$: Xu Yang's work was conducted while he was a final-year UG student at CUHK under Daoyuan Wu's supervision. He is now affiliated with Hong Kong Applied Science and Technology Research Institute (ASTRI).}}
\IEEEauthorblockA{\textit{ASTRI} \\
Hong Kong SAR, China \\
xuyang@astri.org}
\and
\IEEEauthorblockN{Juantao Zhong$^{\ddagger}$\thanks{$^{\ddagger}$: Juantao Zhong is the co-first author.}}
\IEEEauthorblockA{\textit{Lingnan University} \\
Hong Kong SAR, China \\
ericzhong@ln.edu.hk}
\and
\IEEEauthorblockN{Daoyuan Wu$^*$\thanks{$^*$: Daoyuan Wu is the corresponding author.}}
\IEEEauthorblockA{\textit{Lingnan University} \\
Hong Kong SAR, China \\
daoyuanwu@ln.edu.hk}
\and
\IEEEauthorblockN{Xiao Yi}
\IEEEauthorblockA{\textit{The Chinese University of Hong Kong} \\
Hong Kong SAR, China \\
yixiao5428@link.cuhk.edu.hk}
\and
\IEEEauthorblockN{Jimmy H. M. Lee}
\IEEEauthorblockA{\textit{The Chinese University of Hong Kong} \\
Hong Kong SAR, China \\
jlee@cse.cuhk.edu.hk}
\and
\IEEEauthorblockN{Tan Lee}
\IEEEauthorblockA{\textit{The Chinese University of Hong Kong} \\
Hong Kong SAR, China \\
tanlee@ee.cuhk.edu.hk}
\and
\IEEEauthorblockN{Peng Han}
\IEEEauthorblockA{\textit{Lingnan University} \\
Hong Kong SAR, China \\
helenhan@ln.edu.hk}
}
\fi

\maketitle

\begin{abstract}

Online exams conducted via video conferencing platforms such as Zoom have become widespread, yet ensuring exam integrity remains challenging due to the difficulty of monitoring multiple video feeds in real time. We present \name, an online exam proctoring and analysis system that combines lightweight real-time face detection with deep face recognition for post-exam analysis. \name assists invigilators by monitoring student presence during exams and identifies abnormal behaviors, such as face disappearance, face rotation, and identity substitution, from recorded videos. The system addresses three key challenges: (i) \emph{efficient real-time video capture and analysis}, (ii) \emph{automated student identity labeling using enhanced OCR on dynamic Zoom name tags}, and (iii) \emph{resource-efficient training and inference on standard teacher devices}. Extensive experiments show that \name achieves 90.4\% accuracy in real-time face detection and 98.4\% accuracy in post-exam recognition with low overhead, demonstrating its practicality and effectiveness for online exam proctoring.


\end{abstract}

\section{Introduction}
\label{sec:intro}

The rapid adoption of online and hybrid education, accelerated by the COVID-19 pandemic, has fundamentally reshaped teaching and assessment practice worldwide. Synchronous online examinations conducted via video conferencing platforms such as Zoom and Microsoft Teams are now commonplace in schools and universities. While these formates offer flexibility and accessibility, ensuring exam integrity and fairness in remote settings remains a major challenge.
A core difficulty lies in effective exam invigilation. In contrast to traditional in-person exams, where instructors can physically monitor students, online proctoring typically requires supervising dozens of simultaneous video streams displayed as small windows in a gallery view. Manual monitoring under such conditions is inherently limited by human attention, prone to fatigue and bias, and does not scale well for large classes or extended exams. Although commercial automated proctoring systems employ computer vision and AI to address these issues, they often rely on intrusive data collection, centralized cloud processing, or complex installations, raising privacy concerns and limiting deployment in resource-constrained or privacy-sensitive environments~\cite{proctoru_official_website,balash2021examining,balash2023educators}. As a result, many institutions continue to reply on manual video monitoring despite its known shortcomings.

\begin{figure}[!t]
    \begin{adjustbox}{center}
    \includegraphics[trim={8pt 0pt 15pt 0pt},clip,width=.8\linewidth]{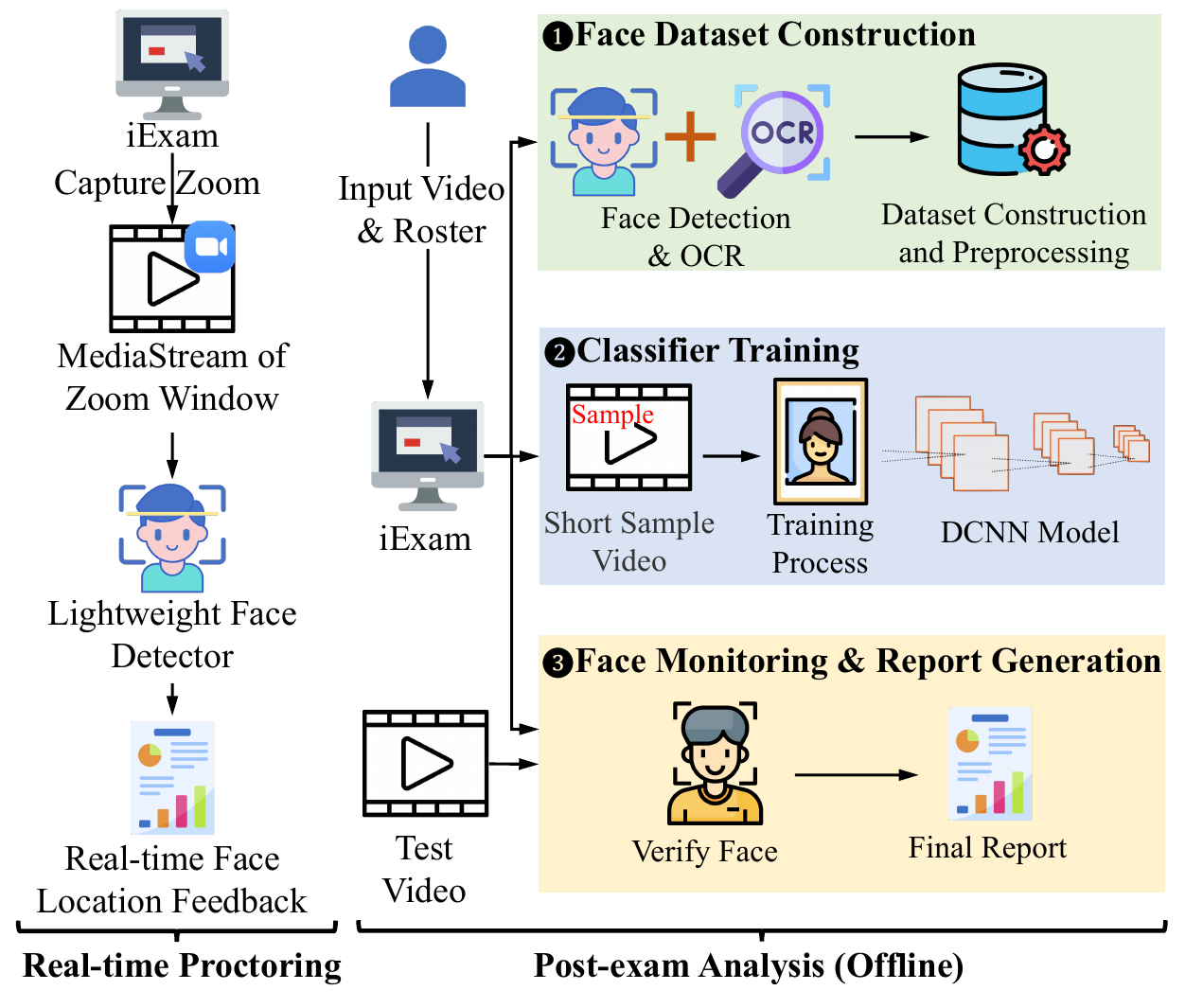}
    \end{adjustbox}
    \caption{The overall workflow of \name, combining lightweight real-time proctoring and deep post-exam analysis.}
    \label{fig:workflow}
\end{figure}

To address these limitations, we present \name, a practical and privacy-preserving online exam proctoring system that combines lightweight real-time face detection with deep learning-based post-exam analysis. The key design insight behind \name is to decouple instant invigilation from in-depth integrity verification, recognizing that these stages impose fundamentally different requirements. During the exam, real-time proctoring demands low latency, scalability and minimal resource consumption. After the exam, retrospective analysis prioritizes accuracy and richer behavioral assessment. As shown in \myfig\ref{fig:workflow}, \name adopts a two-phase architecture. During the exam, a lightweight face detection module runs entirely in the browser capturing the Zoom gallery view. This enables immediate feedback, such as alerts for face absence, while preserving privacy by avoiding the transmission of sensitive video data to external servers. After the exam, recorded video streams are processed offline using Deep Convolutional Neural Networks (DCNNs)~\cite{krizhevsky2012imagenet} for face recognition, enabling robust detection of sophisticated anomalies such as identify substitution or repeated face disappearance. To eliminate manual annotation, \name further introduces an enhanced Optical Character Recognition (OCR)-based labeling pipeline that reliably extracts student identities from dynamic on-screen name labels. To ensure practical deployability on standard teacher laptops, \name incorporates efficiency-oriented optimizations, including sparse OCR labeling and automated model selection, significantly reducing computational overhead without sacrificing accuracy. Extensive experiments on real-world online exam datasets demonstrate that \name achieves 90.4\% accuracy in real-time face detection and 98.4\% accuracy in post-exam recognition, while producing visual reports that reduce instructor workload and improve transparency.

\noindent
\textbf{Contributions.}
In sum, our contributions are follows:
\begin{itemize}
    \item We design and implement \name, a two-stage online exam proctoring system that integrates privacy-preserving real-time face detection and deep learning-based post-exam face recognition and anomaly analysis.
    \item We propose an automated OCR-based labeling pipeline to extract student identities from dynamic Zoom window labels without manual annotation.
    \item We introduce resource-efficient training and inference optimizations that enable effective deployment on standard teacher hardware.
    \item We validate \name on real-world online exam videos, demonstrating high accuracy and actionable reports for scalable and trustworthy exam supervision.
\end{itemize}

\section{Lightweight Real-time Video Monitoring}
\label{sec:realtime}

To enable remote, low-cost, and low-latency exam monitoring with minimal configuration, \name offers a fully online real-time proctoring component optimized for fast and efficient face detection at the edge. This section details the design motivations, key challenges, and implementation details of the lightweight monitoring platform.

\noindent
\textbf{Design Challenges.}
Developing such a system involves three primary challenges:
(\textit{i}) \emph{Screen capture}: Zoom does not provide a public API for direct video stream access, requiring alternative methods to capture exam sessions;
(\textit{ii}) \emph{Detection algorithm selection}: achieving a balance between detection accuracy and the lightweight computation necessary for real-time operation;
(\textit{iii}) \emph{Computation placement}: ensuring that face detection can be performed efficiently and securely, ideally on the client side, to address privacy and network latency concerns.

To address screen capture, we leverage modern browser Web APIs such as \texttt{getDisplayMedia}, allowing users to select and authorize capture of specific application windows (e.g., the Zoom gallery view) as a MediaStream. Our web-based platform thus obtains the live video feed and renders it for in-browser analysis.

For face detection, we offer two JavaScript implementations via OpenCV.js~\cite{opencvjs}:
(1) the classic Haar-cascade method for rapid, lightweight detection, and (2) a modern DCNN detector for higher accuracy.
Both methods can be dynamically loaded and run in the user's browser, enabling flexible trade-offs between speed and performance depending on hardware capabilities.

A key advantage of \name is its privacy-preserving, fully client-side design. All face detection is performed locally within the browser; no video frames or facial features are ever transmitted externally. This edge-based architecture eliminates privacy risks inherent to cloud processing and enables seamless, scalable deployment in real-world educational settings.



\noindent
\textbf{User Workflow.}
During an online exam, invigilators open the web-based \name platform and select the Zoom window to monitor. The system mirrors the chosen window and overlays real-time face detection results, displaying both the original and processed video streams. The face detection model is loaded on demand, and all computation occurs locally, ensuring fast and seamless operation without external dependencies.

This lightweight, privacy-preserving monitoring solution empowers teachers to efficiently supervise online exams, detecting student presence and basic anomalies with minimal setup and no special hardware or server infrastructure.

\section{Comprehensive Post-exam Video Analysis}
\label{sec:postexam}

The post-exam analysis module of \name is designed to automatically detect student absence, face disappearance, and potential identity substitution in recorded exam videos, providing invigilators with concrete and actionable evidence of irregularities.
\name can run cross-platform (e.g., Windows, macOS), and results are delivered through summary reports, including visualizations such as histograms showing face presence over time for each participant.

The workflow of this component consists of three main stages: (1) face dataset construction, (2) classifier training, and (3) face monitoring and feedback generation.
To initiate the process, the proctor provides a short video segment (e.g., five minutes) sampled from the recorded exam session, which is used to train individualized face recognition models for that exam. This training is performed once per exam to adapt to session-specific visual conditions before analyzing the full recording.

To accommodate resource constraints typical on teachers' personal computers, all detected face crops are saved locally as image files and organized by student identity. An adaptive
DCNN,
such as ResNet~\cite{he2016deep}, is then trained to extract robust facial features and enable high-accuracy recognition. Once trained, this model processes the entire exam video, associating detected faces with predicted student names and flagging intervals when a participant is absent or unrecognized.

In this task, two technical challenges must be addressed:
\begin{itemize}
    \item \textit{Automated Labeling:} Each face image in the training dataset must be reliably labeled. However, the dynamic layout of Zoom gallery mode means student locations constantly change, making static position-based labeling infeasible.

    \item \textit{Efficient Model Training and Inference:} The system must be able to rapidly train and test deep learning models on commodity teacher devices, balancing recognition accuracy with limited computational resources.
\end{itemize}

To tackle these, \name leverages computer vision and OCR (optical character recognition) techniques, paired with multi-threaded processing to accelerate dataset construction and model inference.
\myfig\ref{fig:workflow} illustrates the overall workflow.

\subsection{Automatic Ground Truth Collection}
\label{sec:collect}

A major challenge is to automatically obtain labeled face images from the training video, despite students' positions in the Zoom gallery being dynamic. Our solution is as follows:

\noindent
\textbf{Frame Extraction and Face Detection.}
As the video is processed frame by frame, we utilize a pre-trained DCNN-based face detector,
namely Single Shot-Multibox Detector (SSD)~\cite{liu2016ssd} with ResNet-10 backbone~\cite{he2016deep},
resizing extracted face regions to $300 \times 300$ pixels for model compatibility. Alongside face extraction, the (x, y, width, height) of each face is recorded to facilitate subsequent cropping and analysis.

\noindent
\textbf{The Limitation of Static Position Labeling.}
The Zoom gallery mode arranges participants in a $5\times5$ grid, but their order is not fixed. As students join, leave, or lose connection, the positions are shuffled: newcomers are usually placed at the lower right, while others move to fill vacant spots. Thus, it is impossible to rely on a static mapping between cell position and student identity.

\noindent
\textbf{Enhancing OCR for Label Extraction.}
Zoom displays each participant's name in the lower left of their cell, and this label moves together with the video cell as its position changes. We leverage this observation by using OCR to automatically extract names from these label regions. This provides a robust and dynamic way to assign ground truth labels to cropped face images.

\begin{figure}[!t]
    \centering
    \includegraphics[width=0.6\linewidth]{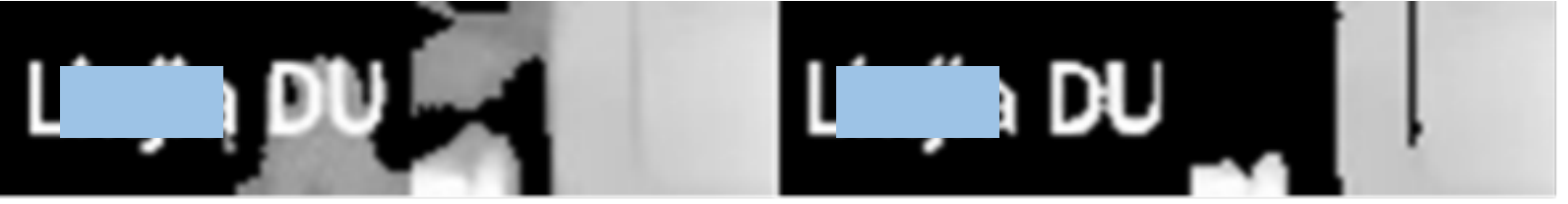}
    \caption{Raw (left) and improved (right) OCR accuracy after thresholding for a participant name.
    }
    \label{fig:thresholdText}
\end{figure}

However, accurate OCR is not trivial: background lighting and low contrast can reduce text recognition accuracy (see left of \myfig\ref{fig:thresholdText}). To address this, we apply adaptive thresholding to enhance contrast between characters and the background, as illustrated in the right of \myfig\ref{fig:thresholdText}. We further enlarge and enhance cropped label regions, running OCR at multiple resolutions to maximize recognition accuracy.

\noindent
\textbf{Post-processing and Dataset Cleaning.} After all candidate face images and labels are collected, the dataset is automatically cleaned. Subfolders (i.e., student classes) with too few images (e.g., less than 100) are treated as noise, since repeated OCR errors for the same person can create multiple small or incorrect subfolders. We also merge folders with similar names (using substring or fuzzy matching with the official roster) to ensure robust label assignment, and prune any folders not corresponding to actual participants. If the video is too short and results in too few ground truth samples, the system prompts the user to provide a longer video for effective training.

\subsection{Effective and Efficient Face Training and Recognition}
\label{sec:recognize}

The second challenge is to efficiently train and deploy robust face recognition models on standard teacher hardware, while ensuring accuracy and fast turnaround.
To address this, we systematically evaluated mainstream face detection and recognition approaches, as well as implemented a series of practical optimizations.

\begin{figure}[!t]
    \begin{adjustbox}{center}
    \includegraphics[trim={30pt 10pt 105pt 43pt}, clip, width=1\linewidth]{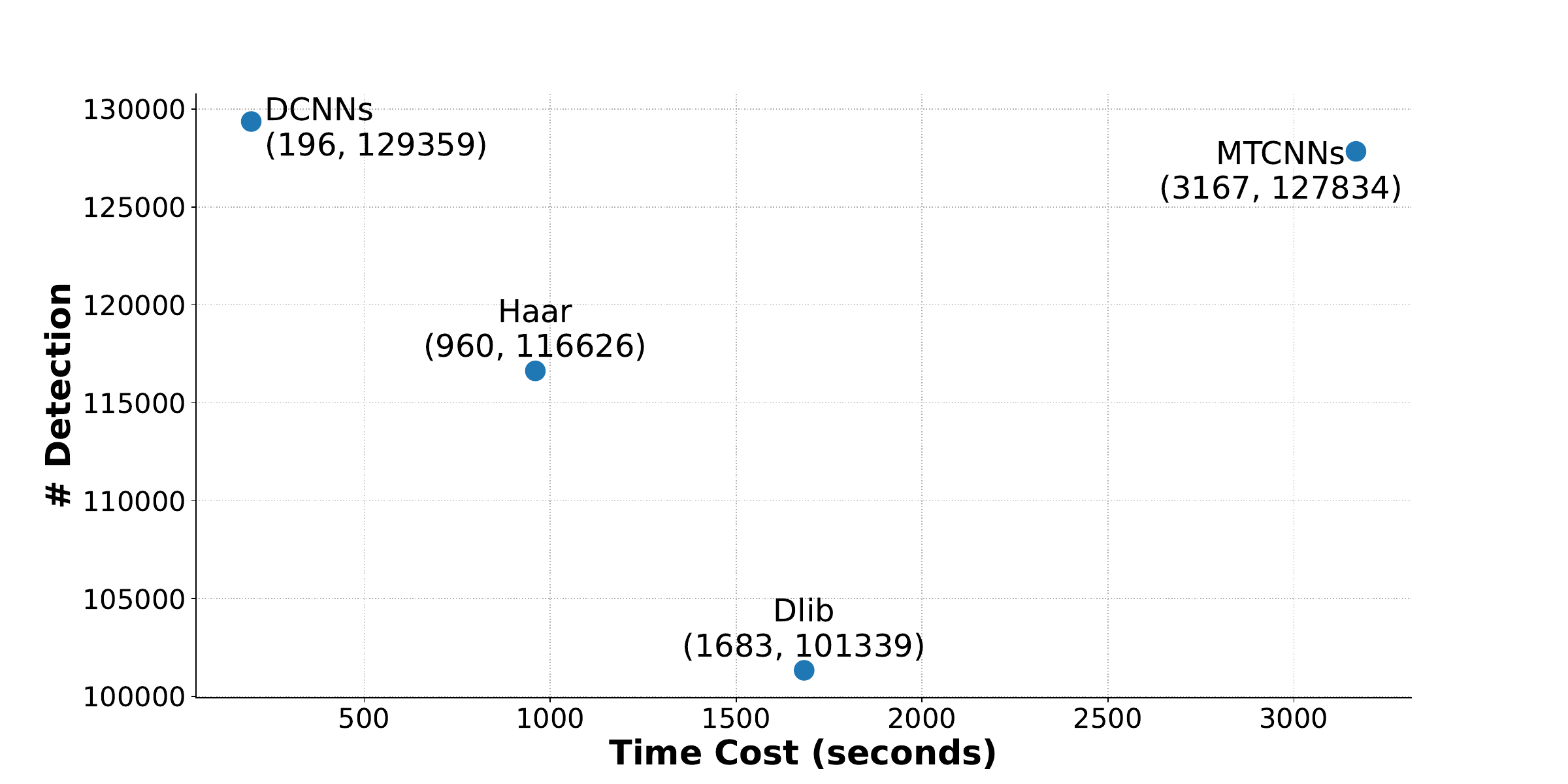}
    \end{adjustbox}
    \caption{Comparison of mainstream face detection algorithms based on a five-minute video.
    }
    \label{fig:detectionComparison}
\end{figure}

\noindent
\textbf{Algorithm Selection and Empirical Comparison.}  
We began by benchmarking several popular face detection methods, including Haar-cascade~\cite{viola2001rapid}, Dlib~\cite{dlib_library}, Multi-task Cascaded Convolutional Networks (MTCNNs)~\cite{zhang2016joint}, and
Deep Convolutional Neural Networks (DCNNs).
As shown in \myfig\ref{fig:detectionComparison}, the DCNN-based detector achieves both the highest detection count (129,359) and the lowest time cost (196 seconds) among all methods evaluated.
MTCNNs and Haar-cascade yield slightly fewer detections, while Dlib records the lowest detection amount.
In terms of efficiency, MTCNNs is by far the slowest (3,167 seconds), followed by Dlib (1,683 seconds) and Haar-cascade (960 seconds).
These results demonstrate that DCNNs not only offers superior detection accuracy but also delivers significantly faster processing, making it the most suitable choice for our real-time and large-scale post-exam analysis requirements.
Based on these findings, we adopted DCNN-based detection
as the foundation for subsequent recognition tasks.

\noindent
\textbf{Model Training and Selection.}  
We evaluated a range of state-of-the-art DCNN architectures, including AlexNet~\cite{krizhevsky2012imagenet}, GoogleNet~\cite{szegedy2015going}, various ResNet variants (ResNet18, ResNet50, ResNet152)~\cite{he2016deep}, as well as SqueezeNet~\cite{SqueezeNet}, MobileNet~\cite{howard2017mobilenets}, and DenseNet121~\cite{huang2017densely}, using cross-entropy loss as the training objective.
To mitigate overfitting arising from the temporal coherence of consecutive video frames, we performed data augmentation such as random flipping, rotation, and cropping during training.
This increases the diversity of training samples and enhances the model's generalization ability.

\begin{figure}[!t]
    \centering
    \includegraphics[trim={40pt 5pt 95pt 65pt}, clip, width=1\linewidth]{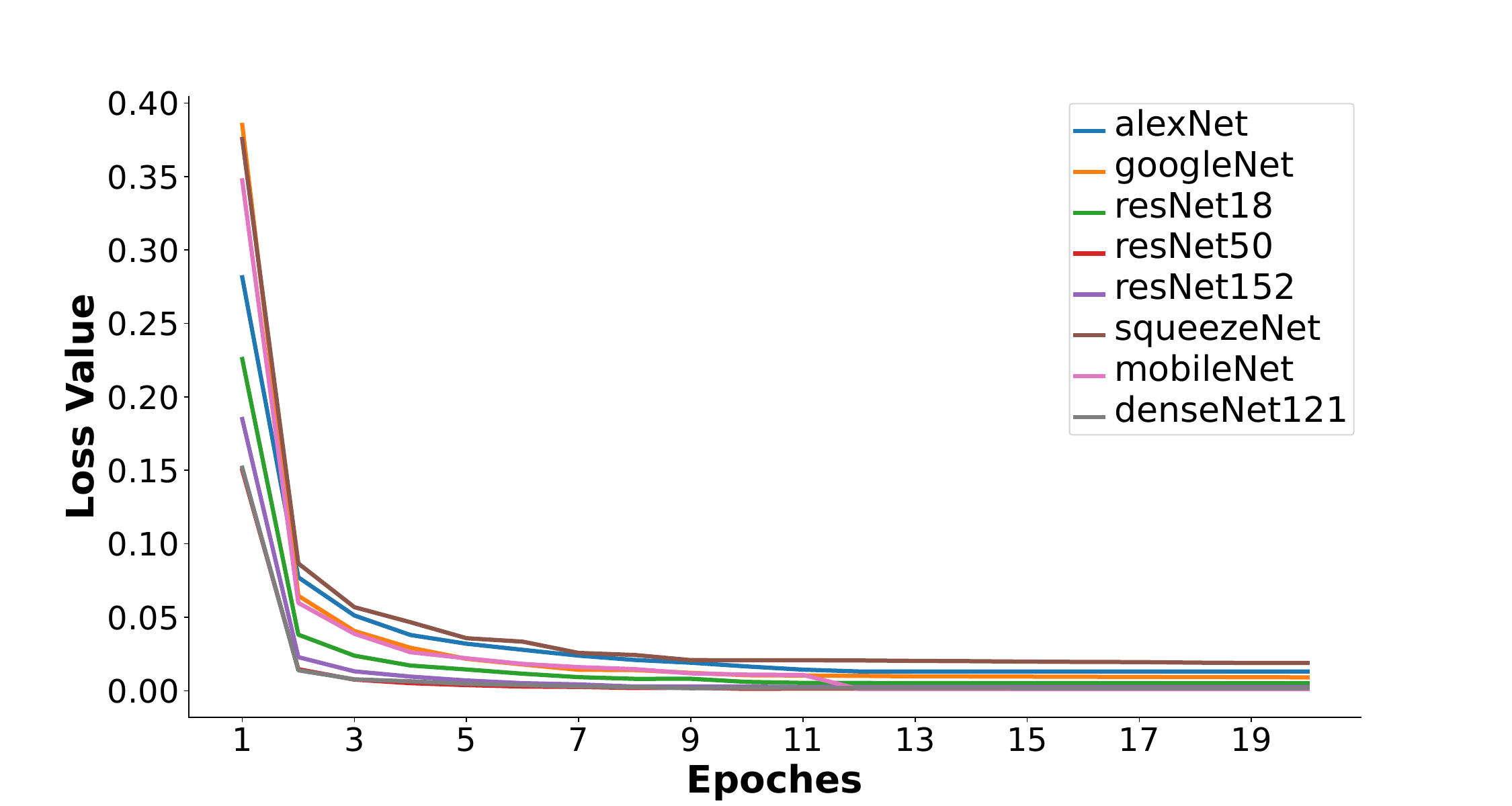}
    \caption{Training loss curves for different DCNN architectures. All models converge rapidly, typically within 10 epochs.
    }
    \label{fig:loss}
\end{figure}

As shown in \myfig\ref{fig:loss}, all tested
DCNN
architectures exhibit rapid convergence, with training loss values dropping sharply within the first few epochs and stabilizing at a low value after approximately 10 epochs.
This indicates that the networks can effectively fit the dataset in a short training period, enabling us to employ early stopping strategies to save computation time without sacrificing accuracy.
In practice, we found that further increasing the number of epochs yielded diminishing returns in loss reduction.
These results demonstrate that mainstream
DCNN
architectures are well-suited to our dataset and application scenario, and our data augmentation strategies successfully prevent overfitting even when large numbers of highly similar images are present.
We ultimately selected the best-performing model, i.e., ResNet50, based on validation accuracy and inference speed in our experiments.

\noindent
\textbf{Pipeline Optimizations.}
To ensure that our face recognition system is both practical and scalable for real-world deployment, we designed a series of optimizations throughout the training and inference pipeline.
These enhancements are aimed at minimizing computation and memory overhead, reducing response latency, and making full use of available hardware resources, all while maintaining high recognition accuracy.
\begin{itemize}
    \item \emph{Sparse OCR Labeling:} To accelerate dataset construction, we perform OCR labeling only once per second for each cell, instead of every frame (as videos typically exceed 20 FPS). The captured name is used as the tag for the entire second, significantly reducing the number of OCR operations and the dataset size without sacrificing accuracy.

    \item \emph{Batch Processing and Multi-threading:} Prior to training, all images are shuffled and loaded in batches. We also increase the number of worker threads for the data loader, and enable multi-threaded operations in all stages to fully exploit CPU parallelism.

    \item \emph{Testing Efficiency:} During inference, predictions are buffered for each position every second, rather than for every frame, which greatly reduces redundant computation. A softmax probability threshold is applied so that only predictions above this value are considered valid, reducing false positives.
\end{itemize}

\noindent
\textbf{Robustness to Recognition Failure.}  
If a student's face is not recognized more than 10 times within a default 30-second interval, the system flags this as a period of disappearance in the report. This design guards against both incidental recognition failures and more severe cases such as proxy attendance or head-turning away from the screen.

Overall, our systematic evaluation and multi-level optimizations allow \name to deliver effective post-exam face recognition, enabling efficient deployment on educational hardware.

\section{Evaluation}
\label{sec:evaluate}

\subsection{The Performance and Accuracy of Face Detection Algorithms in Real-time Exam Monitoring}
\label{sec:testRealTime}

To assess the practical effectiveness of \name's real-time proctoring module, we conducted a comprehensive evaluation of two widely-used face detection algorithms, i,e, Haar-cascade and
DCNNs,
under realistic online exam conditions.

\noindent
\textbf{Experimental Setup.}
We collected data from an actual online exam session, resulting in a two-hour video featuring the Zoom gallery view.  
For this experiment, we implemented two versions of \name's real-time monitoring system: one using the traditional Haar-cascade face detector, and the other employing a modern 
DCNN-based
detector.
Apart from the face detection module, all other components and settings, including browser environment and platform configurations, remained identical.


\begin{figure}[!t]
    \centering
    \begin{subfigure}[t]{.49\columnwidth}
        \includegraphics[width=.95\columnwidth]{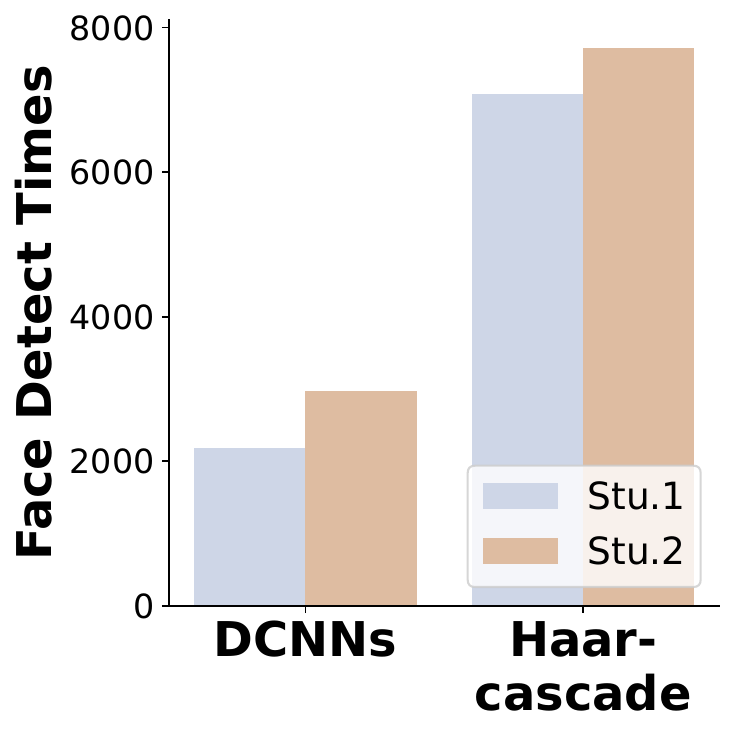}
        \caption{Runtime performance}
        \label{fig:detection_times_comparison}
    \end{subfigure}\hfill
    \begin{subfigure}[t]{.49\columnwidth}
        \includegraphics[width=.95\columnwidth]{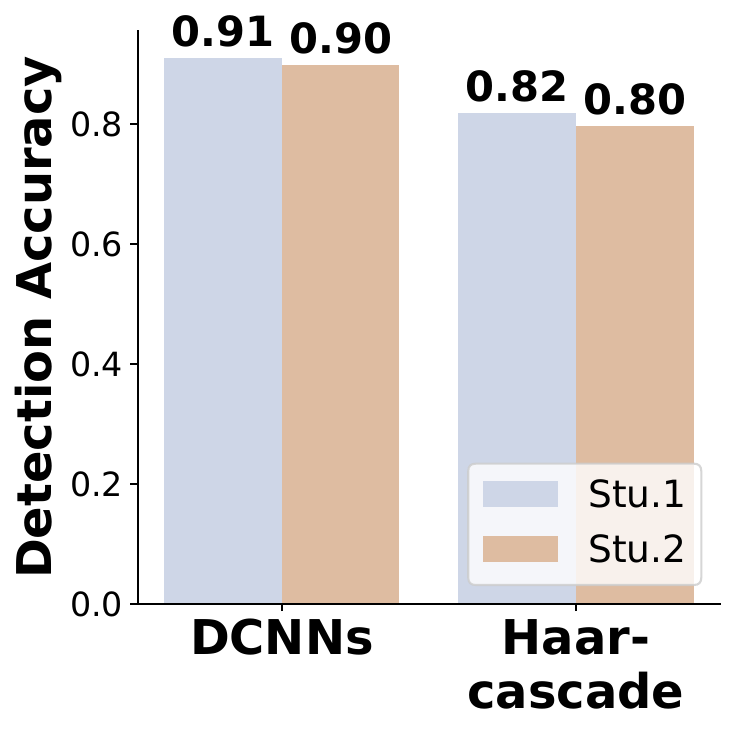}
        \caption{Detection accuracy}
        \label{fig:detection_accuracy_comparison}
    \end{subfigure}
    \caption{Comparison between DCNNs and Haar-cascade face detection algorithms. Face detection times refer to the total number of face recognitions performed by the method over the entire two-hour video.}
    \label{fig:dnn_haar_comparison}
\end{figure}

\noindent
\textbf{Runtime Performance.}
We first measured the
detection efficiency
of each face detection algorithm, focusing on the number of successful detections and the corresponding processing time.
As shown in \myfig\ref{fig:detection_times_comparison}, Haar-cascade achieves about twice the detection rate of
DCNN
on the same video stream, primarily due to its simpler, feature-based approach and lower computational requirements.
However, this higher speed comes at the expense of accuracy and robustness, as the Haar-cascade method is more prone to missing faces under challenging conditions (e.g., large head rotations, partial occlusions).
In contrast, while the
DCNN-based
detector processes frames at a lower rate, requiring roughly twice as long to analyze each image, it consistently provides more reliable detection results, especially in cases where faces are tilted, partially covered, or presented at non-frontal angles.
This robustness is attributed to the learning-based nature of
DCNNs,
which enables them to generalize beyond the fixed feature templates used in traditional methods.


\noindent
\textbf{Detection Accuracy.}
To quantify detection accuracy, we conducted a manual evaluation by randomly sampling 50 frames from the two-hour video and annotating the ground truth for each student's face.
\myfig\ref{fig:detection_accuracy_comparison} illustrates the results, showing that the
DCNN
detector achieves around 90\% detection accuracy, significantly outperforming the Haar-cascade method.
Notably, the
DCNN
approach is better able to adapt to a variety of challenging viewing conditions, such as faces appearing close to the camera, rotated, or partially obscured.
In contrast, Haar-cascade often fails to detect faces that deviate from standard frontal poses or are occluded, and sometimes misidentifies background objects as faces.

These results highlight a fundamental trade-off in real-time video monitoring:
while Haar-cascade is faster, it lacks the robustness and accuracy necessary for reliable online exam proctoring.
The
DCNN-based detector,
despite its relatively higher computational cost, provides the level of precision and adaptability required to maintain exam integrity in practical deployments.
For this reason, \name adopts the
DCNN-based
approach as the default for real-time invigilation.

\subsection{The Performance and Accuracy of Face Recognition Models in Post-exam Video Analysis}

In this section, we present a thorough evaluation of \name's post-exam analysis component, focusing on the training efficiency and recognition accuracy of eight mainstream DCNN-based face recognition models.

\noindent
\textbf{Experimental Setup.}
We first partition the available face image dataset into three disjoint subsets: 70\% for training, 20\% for validation, and the remaining 10\% for testing. All file indices are randomly shuffled prior to splitting to avoid selection bias. To further verify the robustness of our models, we also evaluate on an independent 90-minute exam video not used during training.
All training experiments were performed on a laptop equipped with an NVIDIA GTX 1650Ti GPU (4GB VRAM), simulating a practical deployment environment for educational institutions.

\begin{figure}[!t]
	\begin{adjustbox}{center}
	\includegraphics[trim={60pt 5pt 90pt 3pt}, clip, width=1\linewidth]{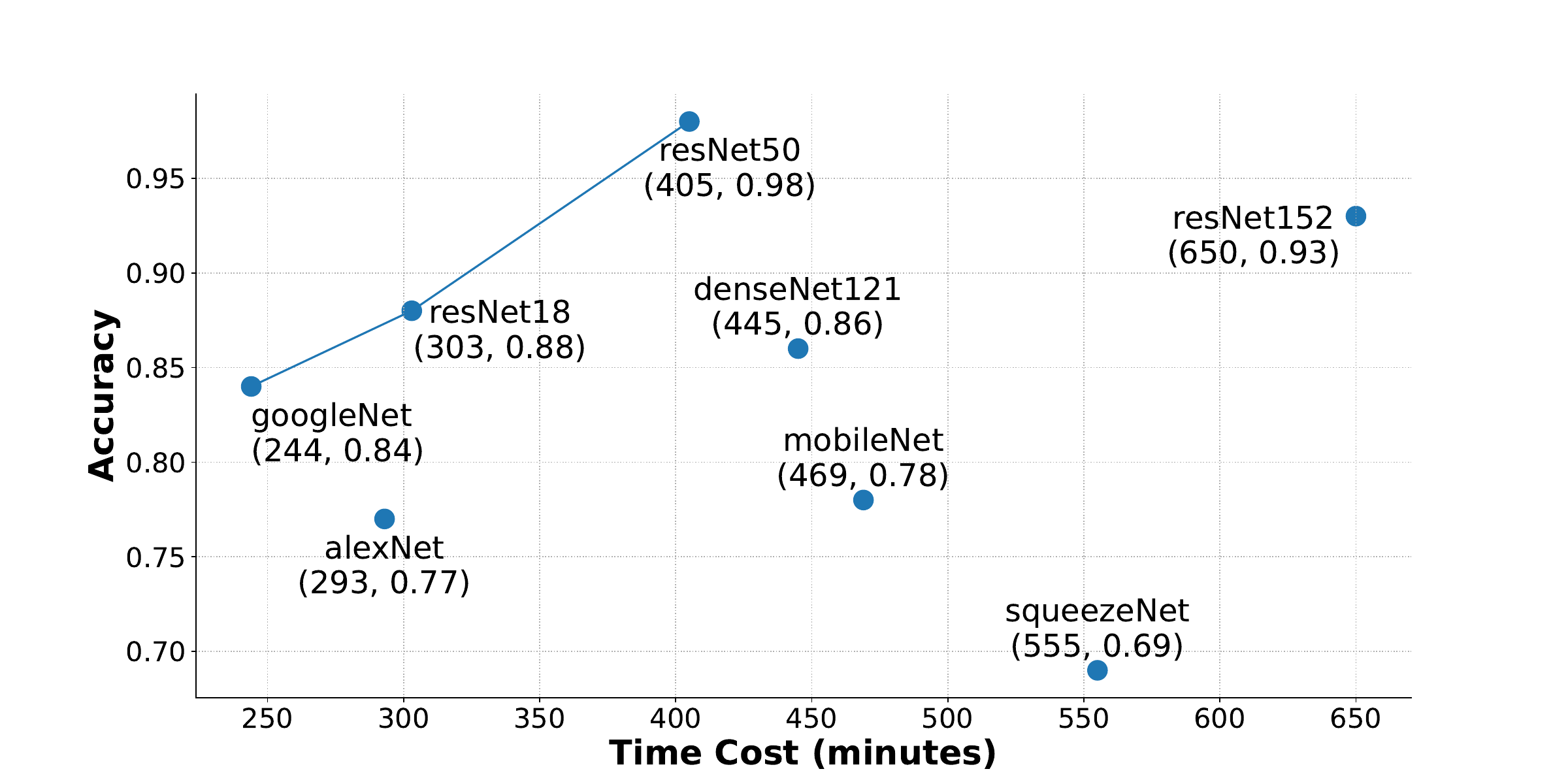}
	\end{adjustbox}
    \caption{Training time cost and recognition accuracy of various DCNN models.
	}
	\label{fig:time}
\end{figure}

\noindent
\textbf{Training Performance.}
We systematically compare the training time and recognition accuracy of eight popular DCNN architectures, including AlexNet, GoogleNet, ResNet18, ResNet50, ResNet152, SqueezeNet, MobileNet, and DenseNet121.
As shown in \myfig\ref{fig:time}, GoogleNet stands out as the fastest model, requiring only 244 minutes to complete 10 epochs of training on our dataset.
AlexNet and ResNet18 also demonstrate relatively fast training speeds, completing in 293 and 303 minutes, respectively.
In contrast, deeper architectures such as ResNet152 and SqueezeNet require substantially more time, 650 and 555 minutes, respectively, reflecting the increased computational complexity of larger models.
MobileNet and DenseNet121 fall in the mid-to-upper range, with training times of 469 and 445 minutes, respectively.

\noindent
\textbf{Recognition Accuracy.}
Despite the differences in training efficiency, we observe notable variation in the face recognition accuracy achieved by each model, as summarized in the \myfig\ref{fig:time}.
ResNet50 achieves the highest accuracy at 98.41\%, establishing a clear lead over other models.
ResNet152 and GoogleNet also perform well, reaching 93.02\% and 88.28\% accuracy, respectively.
AlexNet and MobileNet deliver moderate results, with accuracies of 77.26\% and 78.33\%.
SqueezeNet lags behind with the lowest accuracy at 69.07\%, likely due to its lightweight architecture and reduced parameter count.
DenseNet121 and ResNet18 achieve intermediate performance (86.17\% and 83.82\%, respectively).

This comprehensive evaluation highlights an important trade-off between training time and recognition accuracy:
while models such as GoogleNet and AlexNet train rapidly, their accuracy lags behind deeper architectures such as ResNet50.
Conversely, extremely deep or parameter-heavy networks (e.g., ResNet152, SqueezeNet) incur longer training times without necessarily delivering superior accuracy.
Balancing recognition accuracy and training efficiency is crucial for deployment in real-world online exam scenarios.
Based on our experimental results, ResNet50 emerges as the optimal choice for the post-exam analysis component of \name.
Considering that this component is executed offline, we prioritize accuracy over training time. ResNet50 achieves
state-of-the-art accuracy (98.41\%) while maintaining manageable training time (405 minutes).
This balance makes it well-suited for practical use, ensuring robust and reliable face recognition performance at scale,
without imposing excessive computational burdens on typical teacher or institutional hardware.


\begin{figure*}[!t]
    \centering
    \begin{subfigure}[t]{.67\columnwidth}
        \includegraphics[trim={12pt 10pt 5pt 10pt}, clip, width=\linewidth]{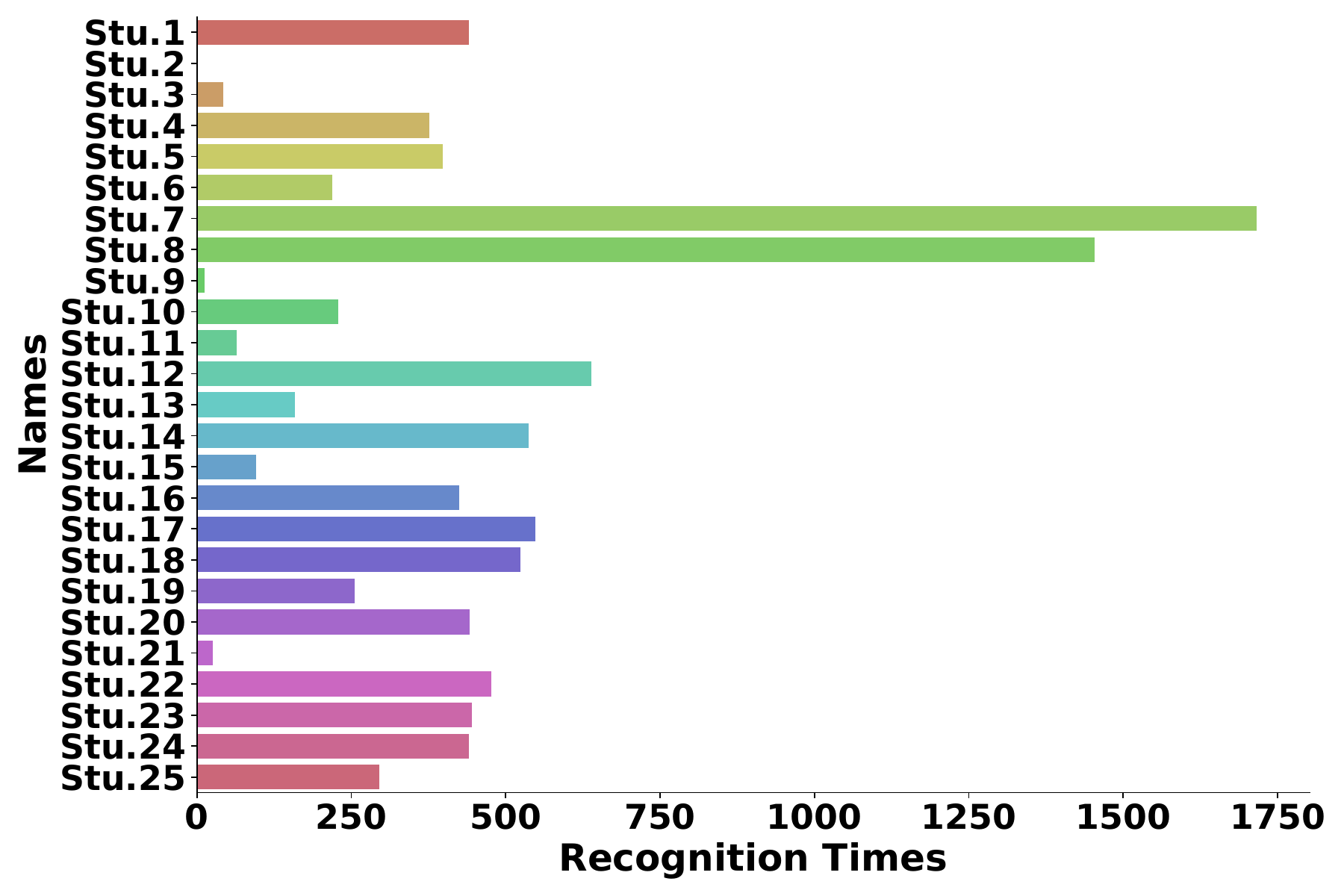}
        \caption{Overall recognition frequency}
        \label{fig:feedback_conclusion}
    \end{subfigure}\hfill
    \begin{subfigure}[t]{.67\columnwidth}
        \includegraphics[width=\linewidth]{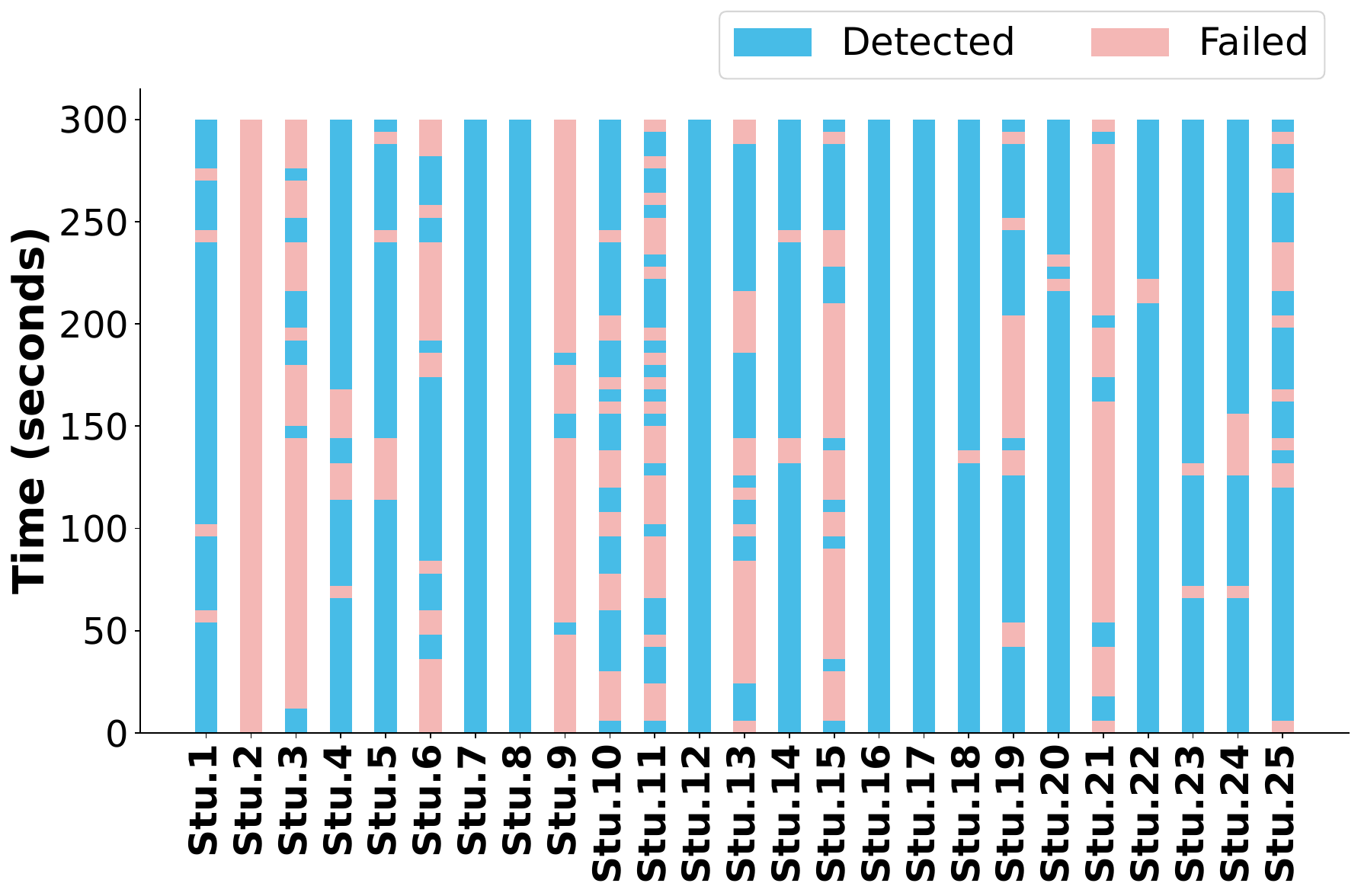}
        \caption{Temporal analysis}
        \label{fig:temporal_analysis}
    \end{subfigure}\hfill
    \begin{subfigure}[t]{.67\columnwidth}
        \includegraphics[trim={5pt 5pt 0pt 15pt}, clip, width=\linewidth]{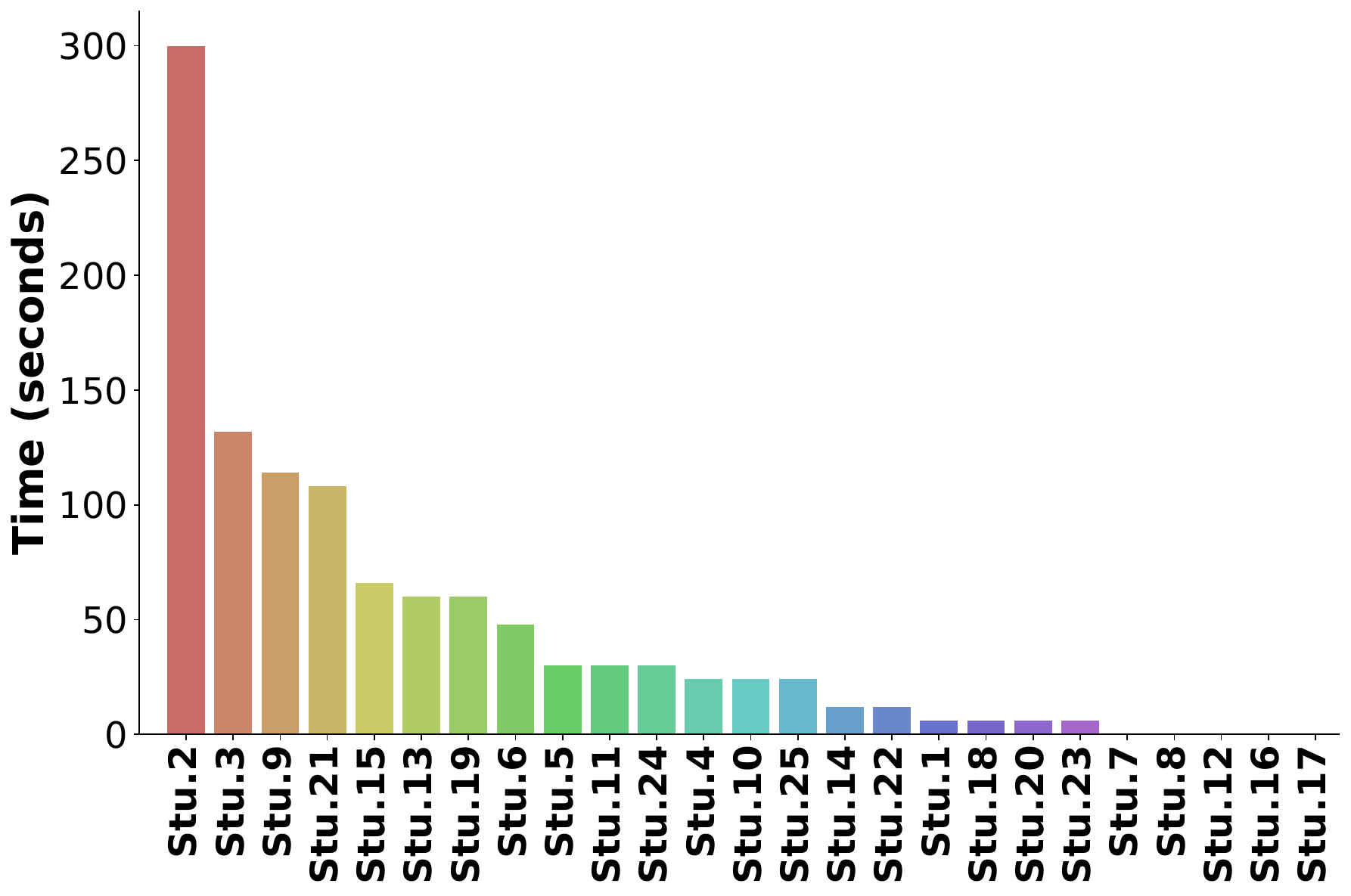}
        \caption{Conclusion of consecutive absence}
        \label{fig:consecutive_absence}
    \end{subfigure}
    \caption{Visual reports generated by \name, showing (a) the number of times each student's face was detected and identified during the exam, (b) a summary of recognition distribution indicating periods when students were \colorbox{CornflowerBlue}{present} or \colorbox{pink}{absent}, and (c) the longest continuous absence interval for each student. Real names have been replaced with generic labels to protect privacy.}
    \label{fig:visual_report}
\end{figure*}

\subsection{A Case Study of \name's Post-exam Analysis Report}
\label{sec:caseStudy}


To demonstrate the practical utility of \name's post-exam analysis, we present a case study using a five-minute exam video. The system automatically generates histogram-style visual reports that allow instructors to quickly review student attendance and detect abnormal behaviors.

\noindent
\textbf{Recognition Frequency Overview.}
As shown in \myfig\ref{fig:feedback_conclusion}, the report includes a bar chart summarizing the total number of times each student's face was recognized. To improve efficiency, one successful recognition per student per second is counted as a valid presence record, rather than counting every frame. This reduces computation while providing a reliable measure of engagement, helping identify students whose faces were rarely detected and potentially flagging technical issues or suspicious behavior.


\noindent
\textbf{Temporal Presence and Absence Analysis.}
The system generates second-by-second presence plots for each participant, as in \myfig\ref{fig:temporal_analysis}. Each segment represents one second of presence or absence, enabling instructors to pinpoint exact times a student disappeared from view. This facilitates cross-checking suspected incidents such as absence or face swapping and provides an auditable record for review.


\noindent
\textbf{Consecutive Absence Detection.}
As in \myfig\ref{fig:consecutive_absence}, \name aggregates consecutive absence durations per student, highlighting extended absences that may require follow-up. Summaries are presented as bar charts and can be cross-referenced with video recordings for validation.


By delivering automated, interpretable, and time-aligned visual feedback, \name's post-exam reports allow instructors to efficiently verify presence, investigate anomalies, and maintain exam integrity, while scaling to longer exams and larger cohorts.


\section{Related Work}
\label{sec:related}

\noindent
\textbf{Deep Learning-based Face Detection and Recognition.} Recent advances in deep learning have substantially improved performance. Convolutional Neural Networks (CNNs)~\cite{howard2017mobilenets,huang2017densely,SqueezeNet,szegedy2015going} and Deep CNNs (DCNNs)~\cite{lecun1998gradient,krizhevsky2012imagenet,he2016deep} achieve high accuracy and robustness, particularly when pre-trained on large-scale face datasets. Several models~\cite{zhang2016joint,wang2023efficientface} demonstrate low detection error rates and fast inference, making them suitable for automated proctoring scenarios.

\noindent
\textbf{Automated Proctoring Systems.}
Recent academic and commercial systems leverage video-based analysis for remote exam monitoring, incorporating face detection, recognition, and anomaly detection~\cite{proctoru_official_website,butler2020systematic}. However, most approaches lack a clear separation between real-time monitoring and post-exam analysis. In contrast, \name adopts a two-stage design that combines lightweight client-side detection with deep offline recognition, explicitly balancing efficiency, privacy, and exam integrity verification.

\section{Conclusion}
\label{sec:conclusion}

This paper presented \name, an effective online exam proctoring system that combines lightweight real-time face detection with deep learning-based post-exam face recognition.
By decoupling instant monitoring from in-depth analysis, \name achieved both efficiency and high accuracy while protecting user privacy.
Experimental results demonstrated that the system delivers robust performance on real-world exam videos, enabling reliable detection of student presence and identification of suspicious behaviors.
Our findings suggested that \name offers a practical and scalable solution for secure and transparent online exam supervision.


\ifANNOYMIZE
\else
\section*{Acknowledgments}
This research/project was partially supported by UGC-funded Courseware Development Grant (ref. no. 4170827) and Funding Scheme for Virtual Teaching and Learning (ref. no. 4170890) from The Chinese University of Hong Kong.
This research has also benefited from financial support of Lingnan University, Hong Kong Special Administrative Region, China.
\fi

\bibliographystyle{IEEEbib}
\bibliography{main}

\end{document}